# Advancing Medical Imaging with Language Models:
## A Journey from N-grams to ChatGPT

Mingzhe Hu[a], Shaoyan Pan[a], Yuheng Li[b] and Xiaofeng Yang[a,b,c*]

[a]Department of Computer Science and Informatics, Emory University, GA, Atlanta, USA
[b]Department of Biomedical Engineering, Emory University, GA, Atlanta, USA
[c]Department of Radiation Oncology, Winship Cancer Institute, School of Medicine, Emory University, GA, Atlanta, USA
*Email: xiaofeng.yang@emory.edu

**Abstract**

In this paper, we aimed to provide a review and tutorial for researchers in the field of medical imaging using language models to improve their tasks at hand. We began by providing an overview of the history and concepts of language models, with a special focus on large language models. We then reviewed the current literature on how language models are being used to improve medical imaging, emphasizing different applications such as image captioning, report generation, report classification, finding extraction, visual question answering, interpretable diagnosis and more for various modalities and organs. The ChatGPT was specially highlighted for researchers to explore more potential applications. We covered the potential benefits of accurate and efficient language models for medical imaging analysis, including improving clinical workflow efficiency, reducing diagnostic errors, and assisting healthcare professionals in providing timely and accurate diagnoses. Overall, our goal was to bridge the gap between language models and medical imaging and inspire new ideas and innovations in this exciting area of research. We hope that this review paper will serve as a useful resource for researchers in this field and encourage further exploration of the possibilities of language models in medical imaging.

## 1. Introduction

In recent years, the healthcare industry has seen significant progress with the integration of artificial intelligence (AI) and machine learning (ML) techniques. Generative Pre-trained Transformer (GPT), with its remarkable language modeling capabilities, has become a popular choice for medical professionals and researchers to analyze and interpret medical data. With OpenAI's recent development of AI chatbot ChatGPT, researchers are interested in using it in healthcare areas such as disease diagnosis, drug discovery, and patient care. Its ability to process large amounts of medical data and provide insights has made it a valuable tool for improving patient outcomes and advancing medical research. The versatility of ChatGPT has made it a go-to solution for many healthcare providers and researchers across different domains.

While prior language models such as BERT (Lee and Toutanova 2018) and ELMo (Sarzynska-Wawer, Wawer et al. 2021) were widely used in the medical field for tasks such as medical image analysis and natural language processing of electronic health records, ChatGPT represents a significant improvement in language modeling. Its ability to generate more human-like, coherent, and contextually relevant responses has made it a valuable tool for medical professionals and researchers in a variety of healthcare applications. ChatGPT's advancements in language modeling have opened up new possibilities for medical diagnosis, patient care, and drug discovery, making it a promising solution for natural language processing tasks in the medical domain.

Despite the wide-ranging applications of language model processing of electronic health records and disease diagnosis, there has been a lack of their applications for medical imaging. Medical image applications, such as disease diagnosis (Lee, Hu et al. 2022, Li, Hsu et al. 2022), multi-organ delineation (using Vnet (Zhou, Rahman Siddiquee et al. 2018), Vision transformer (Pan, Tian et al. 2022), and Multi linear layer Mixer (Pan, Chang et al. 2022, Valanarasu and Patel 2022), lesion detection (Hu, Amason et al. 2021), and high-quality image synthesis (using Generative Adversarial Networks (Lei, Jeong et al. 2018, Lei, Harms et al. 2019, Yamashita and Markov 2020, Pan, Flores et al. 2021) and Denoising Diffusion Probabilistic model (Pan, Wang et al. 2023), requires information which may not be able to be learned from natural language model. As such, it can be challenging to integrate the two modalities effectively. Additionally, the interpretation of medical images typically requires specialized training and can be time-consuming and error-prone. Moreover, there is a lack of review papers that focus on the application of language models in medical imaging analysis, making it difficult for researchers to gain a comprehensive understanding of the field and its potential applications. Given these challenges, researchers have attempted to develop language models that can extract meaningful information from medical images and assist medical professionals in their interpretation. While some progress has been made in this area, much work still needs to be done to realize the full potential of language models for medical imaging analysis.

Therefore, to address the challenges and promote the use of language models in medical imaging, we decided to write this review paper. Our paper aims to serve as a foundational tutorial for researchers in this field, as well as an inspiration for them to innovate and develop new approaches to using language models to improve medical imaging analysis. In this paper, we will begin by providing an overview of the history and concepts of language models, with a special focus on large language models. We will then review the current literature on how language models are being used to improve medical imaging, emphasizing different applications. Figure 1 showcases the breadth of topics covered in our review paper. Finally, we will summarize the current state of the field and discuss potential future directions for research. Our hope is that this review paper will bridge the gap between language models and medical imaging and inspire new ideas and innovations in this exciting area of research.

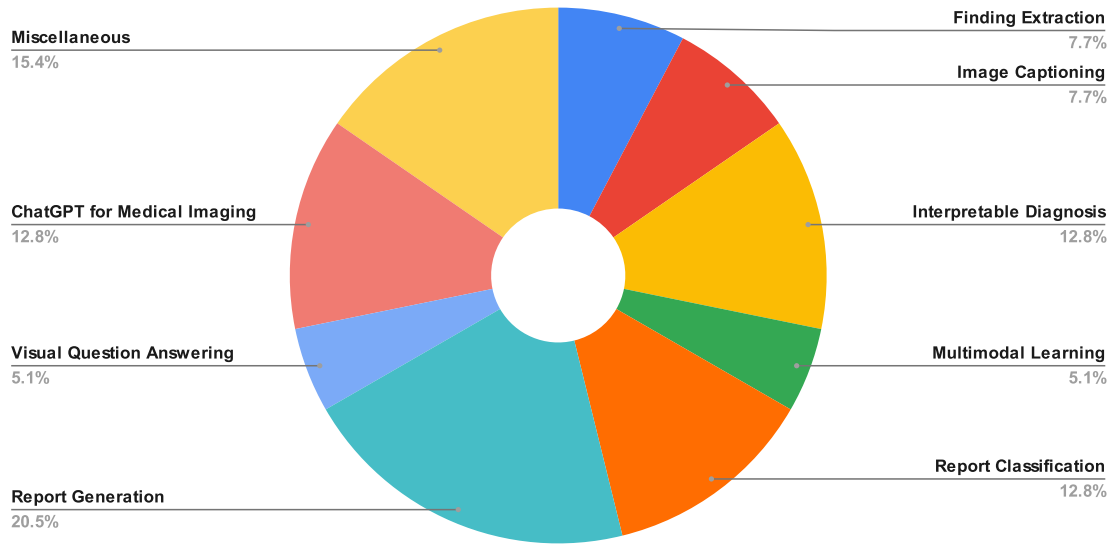

**Figure 1.** An overview of the range of topics that are discussed in our review paper. We can see that report generation is the most common application. Some researchers have begun to apply ChatGPT for medical imaging.

## 2. Basics of Language Models

Language models have a long history dating back to the 1950s, when researchers first began exploring the possibilities of computer-based natural language processing. In the early days, language models were simple statistical models based on n-gram frequencies, which were used to predict the probability of the next word in a sentence based on the previous words. Over time, more sophisticated models were developed, such as hidden Markov models and neural networks, which allowed for more accurate and nuanced language processing. However, it wasn't until the development of large language models, such as GPT (Radford, Narasimhan et al. 2018) and BERT (Lee and Toutanova 2018), that the field truly began to take off. These models use massive amounts of training data and sophisticated algorithms to understand natural language in a way that was previously impossible, opening up new possibilities for a wide range of applications, including in the field of medical imaging. Recent advances in language models include the development of more specialized models for specific tasks, such as medical text processing and dialogue systems, as well as research into novel training techniques, such as unsupervised and few-shot learning. Additionally, there has been increasing interest in developing multilingual and cross-lingual language models that can understand and process multiple languages, which has important implications for healthcare and medical research in diverse linguistic settings.

In this section, we will provide a reader-friendly overview of some popular language models. We will also specifically highlight some powerful large language models that have made significant contributions to natural language processing, such as GPT and BERT. By the end of this section, readers will have a solid understanding of the key concepts and features of these models, setting the stage for a deeper exploration of their applications in medical imaging.

## 2.1 N-grams

N-grams are a sequence of N words used in NLP (Manning and Schutze 1999). The concept is easy to understand and is used for a variety of things in NLP, such as auto-completion of sentences, auto-spell check, and checking for grammar in a sentence. N-gram probabilities are used in NLP models to predict the occurrence of a word in a sentence. To train an NLP model, a large corpus of data is required. Once the model is trained, it calculates the probability of the occurrence of a word after a certain word. In a bigram model, the model learns the occurrence of every two words to determine the probability of a word occurring after a certain word. The probability is calculated by counting the number of times a word occurs in a required sequence, divided by the number of times the word before the expected word occurs in the corpus. The larger the corpus, the better the predictions. NLP and n-grams are used to train voice-based personal assistant bots, and understanding n-grams can help optimize machine learning models.

## 2.2 Recurrent Neural Networks

Recurrent Neural Networks (RNNs) are a type of neural network that allows us to learn from sequential data by considering the order of observations. Unlike feed-forward neural networks, RNNs evolve a hidden state over time, which incorporates information from the previous hidden state and the latest input(Rumelhart, Hinton et al. 1985). This enables us to chain a sequence of events together and perform backpropagation through time. RNNs can handle arbitrary length inputs and outputs, making them versatile for a broad range of sequence learning tasks. One to many, many to one, many to many (same), and many to many (different) are the general architectures used for various sequence learning tasks. RNNs are particularly useful in natural language processing and speech recognition, where the order of observations is critical for accurate analysis. RNNs have been successfully used in many applications, such as language modeling, machine translation, speech recognition, and image captioning. However, they suffer from some limitations, such as the vanishing gradient problem, which affects the ability of the network to capture long-term dependencies. Therefore, several variants of RNNs have been developed, such as Long Short-Term Memory (LSTM) and Gated Recurrent Unit (GRU).

## 2.3 Long Short-Term Memory

LSTM (Hochreiter and Schmidhuber 1997) is a type of RNN that excels at processing sequential data with long-term dependencies. Unlike traditional RNNs, LSTM can address the vanishing gradient problem, which arises when gradients become too small as they propagate through the network, making it difficult for the network to learn long-term dependencies. LSTM introduces a memory cell and several gates that control the flow of information into and out of the cell. The memory cell stores information over time, while the gates decide what information to keep or discard at each time step. These gates, namely the forget gate, input gate, and output gate, respectively control the amount of old information to forget, the amount of new information to store, and the amount of information to output. By selectively remembering or forgetting information over long periods of time, LSTM is highly suited for tasks like speech recognition, language translation, and image captioning. Overall, LSTM is a significant improvement over traditional RNNs and has revolutionized the field of deep learning by enabling the efficient processing and analysis of sequential data.

## 2.4 Transformers

Transformers (Vaswani, Shazeer et al. 2017) for natural language processing is a revolutionary model architecture that has outperformed traditional encoder-decoder models based on recurrent or convolutional neural networks. The introduction of attention mechanisms optimized the performance of encoder-decoder models. With the help of attention mechanisms, Transformer, the first model that entirely relies on self-attention mechanisms to calculate input and output representations, has outperformed recurrent and convolutional neural networks. The Transformer model can be trained in parallel, which reduces training time. The encoder consists of six identical blocks, each of which has two sub-layers, a multi-head self-attention mechanism, and a fully connected feed-forward network. The decoder is also made up of six identical blocks, each with three sub-layers, which includes the addition of a multi-head self-attention sub-layer. The Transformer model's self-attention mechanism utilizes a scaled dot-product attention approach, which multiplies the query and key vectors and applies a softmax function to the results to obtain the weightings of the output. While both additive and dot-product attention are common attention functions, dot-product attention is more space-efficient and faster because it can use a highly optimized matrix multiplication code to implement.

## 3. Large Language Models

Large Language Models (LLMs) are neural network-based models widely used in NLP tasks, capable of generating human-like text by predicting the next word in a given sentence or generating new sentences. These models are trained on massive amounts of text data, allowing them to learn the underlying patterns and structures of natural language, resulting in the more coherent and contextually relevant text. Additionally, LLMs can capture long-term dependencies in language, generating more coherent and cohesive text. Their ability to perform various language-related tasks, such as text classification, sentiment analysis, and text generation, make them valuable tools in different NLP applications, including medical image analysis. In conclusion, LLMs represent a significant breakthrough in NLP, and their ability to generate high-quality, contextually relevant text makes them an essential tool for researchers and practitioners alike.

## 3.1 BERT

Bidirectional Encoder Representation from Transformers (BERT) (Lee and Toutanova 2018) is a pre-trained LLM that has gained widespread attention in the NLP community due to its exceptional performance on a range of tasks. Unlike previous LLMs that relied solely on unidirectional training, BERT uses a bidirectional training approach, where it learns to predict masked words in a sentence by taking into account both the left and right contexts. This allows the model to capture more complex relationships between words and phrases and achieve a better understanding of the underlying meaning of the text. Furthermore, BERT's pre-training process involves training on a large and diverse corpus of text, which allows it to learn the nuances and intricacies of natural language. This has led to BERT achieving state-of-the-art performance on several benchmarks, including sentiment analysis, question-answering, and natural language inference. Overall, BERT has revolutionized the field of NLP and paved the way for more advanced language models.

### 3.2 paLM

Parameterized Language Model (PaLM) (Peng, Schwartz et al. 2019) is a large language model that was proposed in 2019. Unlike traditional language models that use a fixed number of parameters to generate text, paLM employs a dynamic parameterization approach. This means that the model can adapt its parameters based on the context and input, allowing it to generate more accurate and relevant text. PaLM is trained on large amounts of text data using unsupervised learning techniques, which enables it to learn the underlying patterns and structures of natural language. It is a versatile model that can be fine-tuned for various NLP tasks, including text classification, question answering, and language modeling. One of the key features of paLM is its ability to handle out-of-vocabulary (OOV) words, which are words that are not present in the training data. The model can effectively generate contextually relevant words to replace OOV words, improving its overall text generation performance.

### 3.3 LLaMA

Large Language Model Meta AI (LLaMA) (Touvron, Lavril et al. 2023) is a collection of foundation language models ranging from 7B to 65B parameters. These models were trained on trillions of tokens using unsupervised learning techniques such as masked language modeling and next-sentence prediction. The training data was sourced from publicly available datasets such as Wikipedia, Common Crawl, and OpenWebText. By training exclusively on publicly available data, the LLaMA team has shown that it is possible to achieve state-of-the-art performance without relying on proprietary datasets.

### 3.4 ChatGPT

ChatGPT is a large language model based on the GPT architecture (Radford, Narasimhan et al. 2018), developed by OpenAI. It was first introduced in June 2020, and several versions of the model have been released. The model was trained on a massive amount of text data, including books, articles, and web pages, allowing it to learn the nuances and intricacies of natural language.

One of the unique features of ChatGPT is its ability to generate human-like text in response to a given prompt. This makes it useful for a wide range of natural language processing tasks, including chatbots, language translation, and text summarization, among others. ChatGPT can also be fine-tuned on specific tasks by providing it with a smaller, task-specific dataset.

The original version of ChatGPT, GPT-1, was trained on a dataset of 40GB of text data, while later versions, such as GPT-2 and GPT-3, GPT-4 were trained on significantly larger datasets. This increase in the amount of training data has allowed ChatGPT to improve its performance significantly, with GPT-4 being capable of generating text that is almost indistinguishable from that written by a human.

## 4. Language Models for Medical Imaging
4.1 Finding Extraction from Radiology Report

Extracting meaningful information from radiology reports is essential for secondary applications in clinical decision-making, research, and outcomes prediction. However, there remain several challenges in the field, such as reducing the workload of radiologists and improving communication with referring physicians. Additionally, extracting comprehensive semantic representations of radiological findings from reports takes a lot of work.

In this section, we will discuss the current state of the art in finding extraction from radiology reports, including the challenges and limitations of existing methods and the potential benefits of novel frameworks and techniques. By exploring these advancements, we hope to provide valuable insights into using language models in improving clinical decision-making in radiology.

One promising approach is ChestXRayBERT, a framework proposed by (Cai, Liu et al. 2021) that uses a pre-trained BERT-based language model to generate the "impression" section of chest radiology reports automatically. This approach significantly reduces the workload of radiologists and improves communication between radiologists and referring physicians. In experiments, ChestXRayBERT outperformed existing state-of-the-art models in terms of readability, factual correctness, informativeness, and redundancy, achieving an average score of 4.23 out of 5.

Another approach for extracting valuable information from radiology reports is proposed by (Lau, Lybarger et al. 2022) who developed a corpus of radiology reports annotated with clinical findings using an event-based schema. They then used this annotated corpus to train a pre-trained language model called BioBERT to extract clinical findings from radiology reports. Results show that BioBERT achieved an overall F1 score of 95.6% for triggers, 79.1% for span-only arguments, and 89.7% for span-with-value arguments.

Liu *et al.* (Liu, Zhang et al. 2022) aimed to improve clinical decision-making in radiology by developing an ensemble learning classification model using NLP applied to the Chinese free text of radiological reports to determine their value for liver lesion detection in patients with colorectal cancer. They found that the traditional Tf-idf statistical data model outperformed the word2vec semantic method in structured report classification, achieving an F1 value of 0.98. This study provides valuable insights into the use of NLP and ensemble learning models for improving clinical value and resource utilization in health care management.

*Assessment*

Although language models for finding extractions from radiology reports have shown promising results, there are still some limitations that need to be addressed. One limitation is the need for more generalizability to diverse datasets, as many models have only been evaluated on limited datasets. Another limitation is the need for manual annotation, which can be time-consuming and potentially prone to errors. Additionally, some models do not consider the context of the report outside of the "findings" section, which may limit their ability to generate summaries that are consistent with the overall report.

From the future perspective, there are opportunities for improvement in language models for finding extractions from radiology reports. One suggestion is to explore more advanced natural language processing techniques to improve the accuracy and efficiency of clinical finding extraction. Another suggestion is to expand the corpus to include more diverse patient populations and imaging modalities to improve generalizability. Standardized reporting using tools such as the Liver Imaging Reporting and Data System can also reduce variability in reporting and improve model performance. Overall, future research should continue to explore the potential of language models in improving clinical decision-making in radiology. A summary of the works we reviewed in this section is given in Table 1.

**Table 1.** Overview of language models for finding extractions.

| References | ROI | Modality | Vision Model | Language Model |
|---|---|---|---|---|
| (Cai, Liu et al. 2021) | Chest | X-ray | None | ChestXRayBERT |
| (Lau, Lybarger et al. 2022) | General | CT, X-ray | None | BioBERT |
| (Liu, Zhang et al. 2022) | Liver | CT, X-ray | None | N-gram |

4.2 Image/Video Captioning

Generating accurate and reliable automatic report generation systems for medical images poses several challenges, including analyzing limited medical images using machine learning approaches and generating informative captions for images involving multiple organs. Captioning fetal ultrasound images is a particularly challenging task due to the complexity and variability of the images. Fetal ultrasound images are often noisy, low-resolution, and can vary greatly depending on factors such as fetal position, gestational age, and imaging plane. Additionally, there needs to be large-scale annotated datasets for this task. To address these challenges, researchers have proposed deep learning-based approaches that integrate visual and textual information to generate informative captions for images and videos.

Generating captions for limited CT and digital breast tomosynthesis (DBT) images is a challenging task due to the complexity and limited information present in the images. (Alsharid, Cai et al. 2022) presented a novel approach to generate medical captions for limited CT and DBT images using a multilevel transfer learning technique and LSTM framework. The study highlights the challenges of generating captions for medical images involving multiple organs and the difficulty of analyzing limited medical images using machine-learning approaches. The authors propose a deep recurrent architecture that combines Multi Level Transfer Learning (MLTL) framework with a LSTM model. The proposed approach achieved an accuracy of 96.90% and a BLEU score of 76.9%, which outperforms existing methods in generating accurate and informative captions for limited CT and DBT images.

In another study, (Alsharid, El-Bouri et al. 2021) proposed a course-focused dual curriculum method for captioning fetal ultrasound images. The method trains a model to dynamically transition between two different modalities (image and text) as training progresses as shown in figure 2. Two configurations of the course-focused dual curriculum are compared, and the best results are achieved with a curriculum that focuses on image captioning in the early stages of training and then transitions to text-based captioning. The proposed method outperforms existing methods on the fetal ultrasound dataset. The authors suggest that their approach could be applied to other types of medical imaging as well.

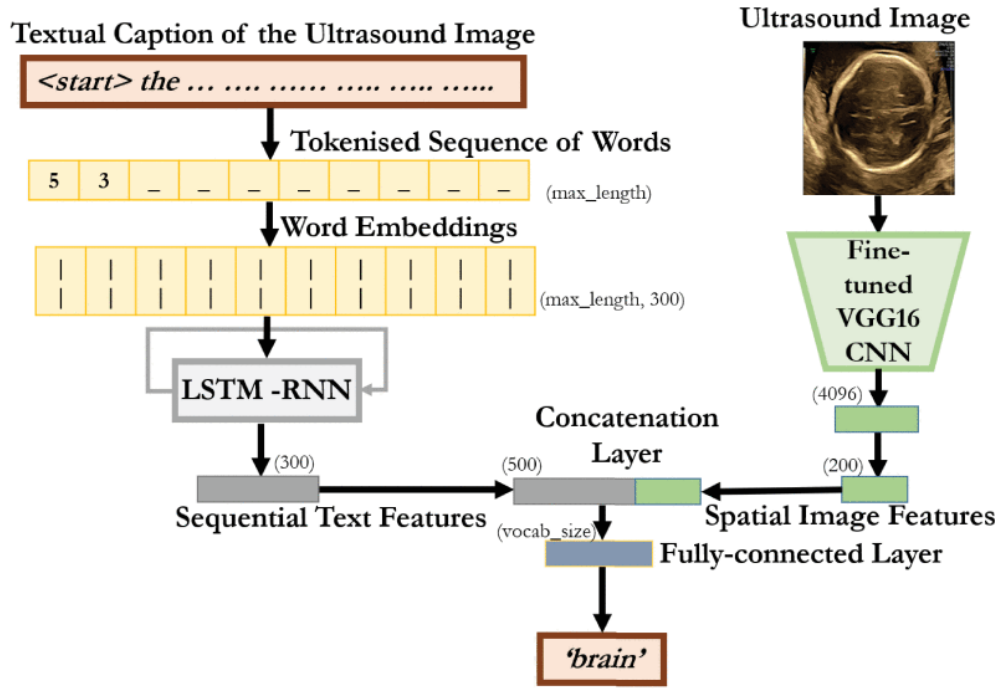

**Figure 2.** Courtesy of (Alsharid, El-Bouri et al. 2021). The model comprises two branches: the right branch processes image information using a fine-tuned VGG-16 and fully-connected layers, while the left branch embeds tokenized words using a Word2vec embedding vector and passes them through a recurrent neural network. The outputs of both branches are concatenated and used to predict the next word in the captioning sequence.

Caption Generation cannot be only applied to images but also videos. In the author's another work, (Aswiga and Shanthi 2022) proposed a novel approach for automatic captioning of fetal ultrasound videos using a three-way multi-modal deep neural network. The study addresses the challenge of generating informative captions for fetal ultrasound videos, which can assist clinicians in their diagnosis and treatment decisions. The proposed method integrates visual, textual, and gaze information to generate captions. The model was trained on a large dataset of fetal ultrasound videos and achieved state-of-the-art performance. The results demonstrate the potential of this technology in improving the accuracy and efficiency of fetal ultrasound diagnosis.

*Assessment*

While recent studies have made significant progress in medical image and video captioning using language models, there are still several limitations that need to be addressed. One major challenge is the small size of the datasets used in some studies, which can affect the generalizability of the proposed approach to larger datasets. Additionally, evaluation metrics used in some studies may not fully capture the quality of generated captions. Future research can explore more sophisticated evaluation metrics to better assess the quality of generated captions.

Another area for improvement is the requirement of large amounts of annotated data for training, which can be difficult to obtain in medical imaging applications. To overcome this, future work could explore ways to reduce the reliance on pre-trained models and to improve the generalization of the method. In addition, incorporating additional sources of information, such as clinical metadata or patient history, into the captioning process could provide more context and improve the accuracy and usefulness of the generated captions.

To improve the performance of the proposed methods, future research could also explore more sophisticated transformer-based word embedding models and spatio-temporal visual and textual feature extractors. Using larger and more diverse datasets can also help address limitations such as lack of diversity in the fetal ultrasound videos. These improvements can ultimately lead to more accurate and reliable automatic report generation systems in the medical field. A summary of the works we reviewed in this section is given in Table 2.

**Table 2.** Overview of language models for image/video captioning.

| References | ROI | Modality | Vision Model | Language Model |
|---|---|---|---|---|
| (Aswiga and Shanthi 2022) | Breast | CT, DBT | MLTL | LSTM |
| (Alsharid, El-Bouri et al. 2021) | Fetal | Ultrasound | VGG16 | LSTM-RNN |
| (Alsharid, Cai et al. 2022) | Fetal | Ultrasound | VGG16 | LSTM, Word2vec |

### 4.3 Diagnosis Interpretability

Medical image analysis and diagnosis have long been challenging due to the need for more interpretability of deep neural networks, limited annotated data, and complex biomarker information. These challenges have inspired researchers to develop innovative AI-based methods that can automate diagnostic reasoning, provide interpretable predictions, and answer medical questions based on raw images.

One approach that emphasizes interpretability is the encoder-decoder framework used by (Zhao, Tian et al. 2021) in their automatic report generation model for thyroid nodules in ultrasound images. By combining medical features with deep network features, the model offers more interpretable reasons to support decision-making, achieving a 95% description accuracy on clinical ultrasound images.

(Monajatipoor, Rouhsedaghat et al. 2022) proposed a different approach to interpretability using BERTHop, a vision-and-language model for chest X-ray disease diagnosis. Their transformer-based model captures associations between medical images and associated text, and PixelHop++ extracts visual representations from X-ray images as shown in figure 3. With an AUC-ROC score of 0.862 on the ChestX-ray14 dataset, BERTHop outperforms state-of-the-art models, reducing medical mistakes in disease diagnosis.

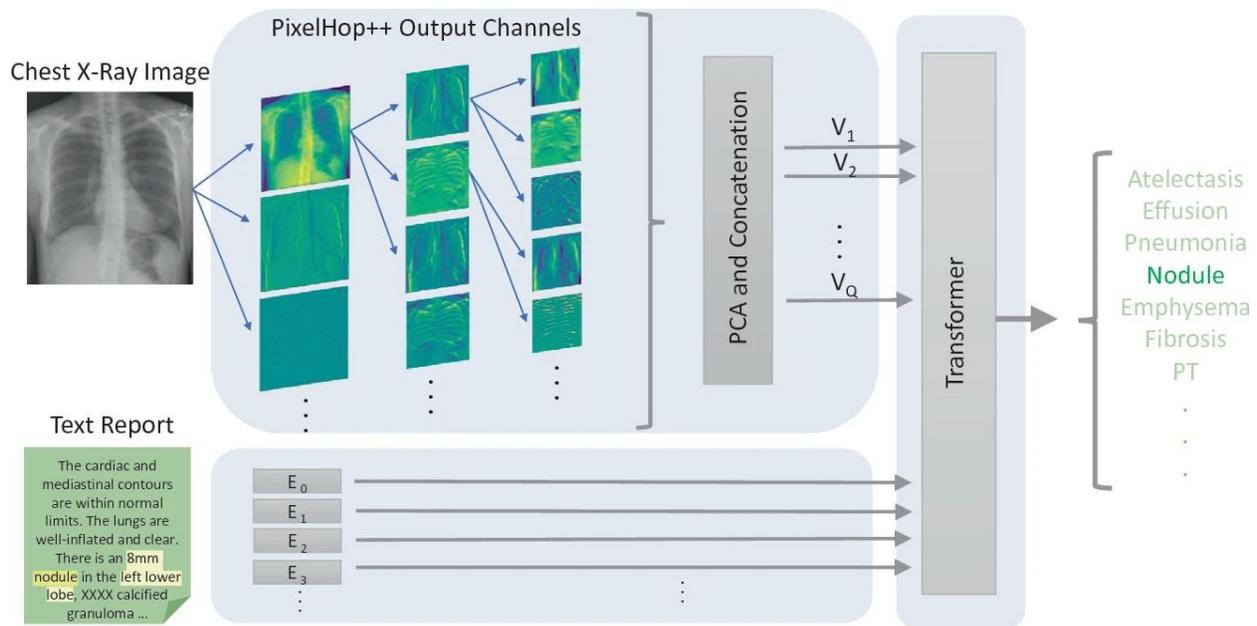

**Figure 3.** Courtesy of (Monajatipoor, Rouhsedaghat et al. 2022). Overview of the proposed approach for modeling relationships between visual and textual modalities in X-ray image analysis. The approach involves three main steps: (1) visual feature extraction using PixelHop++, (2) text encoding using subword embeddings, and (3) joint transformer modeling to capture implicit alignments between the visual and textual modalities.

Visually interpretable diagnosis is also essential. (Zhang, Chen et al. 2019) presented a pathology whole-slide diagnosis method that uses AI to address the lack of interpretable diagnosis in cancer pathology. Their deep convolutional neural network automates human-like diagnostic reasoning and translates gigapixels directly into interpretable predictions, achieving an accuracy rate of 90.6% on urothelial carcinoma diagnosis. They also propose a novel medical image diagnosis network called MDNet that provides semantically and visually interpretable diagnoses, with an accuracy of 0.78 on the ChestX-ray14 dataset for multi-label classification of thoracic diseases.

More advanced transformer-based structures have also been adopted. (Naseem, Khushi et al. 2022) introduced a novel Vision-Language Transformer for Interpretable Pathology Visual Question Answering (PathVQA), which aims to answer medical questions based on pathology images. Their transformer-based architecture combines visual and textual information to embed vision and language features, achieving an accuracy of 70.2% on the PathVQA dataset. This approach presents a promising avenue for medical question-answering that could have significant implications for healthcare.

*Assessment*

Despite the promising results demonstrated by the studies mentioned above, there are still some limitations and open questions that need to be addressed for the development of more interpretable language models in medical imaging. One common limitation is the need for larger and more diverse datasets to ensure the generalizability of the proposed models to other medical domains and tasks. Another challenge is to improve the interpretability of the generated reports, which requires more explicit reasoning steps and finer attention localization. Future research could also explore extending these models to other medical modalities and incorporating more advanced visualization techniques to improve interpretability. Despite these limitations, the development of language models for interpretable diagnosis holds great potential for improving healthcare outcomes and reducing diagnostic errors. A summary of the works we reviewed in this section is given in Table 3.

**Table 3.** Overview of language models for improving the diagnosis interpretability.

| References | ROI | Modality | Vision Model | Language Model |
|---|---|---|---|---|
| (Zhao, Tian et al. 2021) | Thyroid | Ultrasound | Unet, VGG16 | RNN |
| (Monajatipoor, Rouhsedaghat et al. 2022) | Chest | X-ray | PixelHop ++ | BlueBERT |
| (Zhang, Chen et al. 2019) | Bladder | Histopathology | CNN | LSTM |
| (Naseem, Khushi et al. 2022) | General | Histopathology | ResNet-152 | BioELMo |
| (Zhang, Xie et al. 2017) | Bladder | Histopathology | ResNet | LSTM |

### 4.4 Report Classification

In this session, we will be discussing the challenges faced in report classification for medical imaging. The manual labeling process of radiology reports for computer vision applications is time-consuming and labor-intensive, creating a bottleneck in model development. The goal of this research area is to automate this process, improving efficiency and accuracy in the medical field. Extracting clinical information from these reports is also a challenge, limiting the efficiency and accuracy of clinical decision-making. We will explore the use of deep learning-based natural language processing techniques to accurately classify medical imaging reports, addressing the limitations of traditional machine learning methods.

One study by (Wood, Kafiabadi et al. 2022) aimed to develop a deep learning model to automate the labelling of head MRI datasets for computer vision applications. The authors used type-token ratio and Yules I to calculate the linguistic complexity of their report corpus and compared it to similar-sized head CT and chest radiograph corpora from the radiology literature. The deep learning model achieved an accuracy of 0.94 in assigning labels to corresponding examinations, which is comparable to human performance. The introduction of the attention mechanism can further enhance the classification ability of the model. In another study, (Wood, Lynch et al. 2020) presented Automated Labelling using an Attention model for Radiology reports of MRI scans (ALARM). The authors modified and fine-tuned the state-of-the-art BioBERT language model to create ALARM, which uses attention mechanisms to accurately label reports. The results showed that ALARM's classification performance was only marginally inferior to an experienced neuroradiologist for granular classification, suggesting that ALARM can feasibly be used in real-world applications.

Several studies have demonstrated the usefulness of pre-trained language models for automating medical imaging tasks. Pre-training on large corpora of medical text can improve performance in tasks such as report classification. In their study, (Bressem, Adams et al. 2020) aimed to develop a highly accurate classification system for chest radiographic reports using a deep learning natural language model pre-trained on 3.8 million text reports. The results showed that the model achieved an accuracy of 90.5% in classifying chest radiographic reports, which is significantly higher than previous studies. Zachary Huemann aimed to develop and evaluate a NLP models for automated extraction of Deauville scores (DS) from PET/CT reports. The study used multiple pre-trained language models, including BERT, RoBERTa, and ALBERT, which were adapted to the nuclear medicine domain using masked language modeling. The best NLP model achieved an accuracy of 0.91 in extracting DS from the reports, with an F1 score of 0.89.

By comparing changes in report classification over different time periods, it is possible to reveal subtle trends that may otherwise go unnoticed. (Min, Xu et al. 2021) aimed to assess the changes in the acuity of brain MRI findings during the early COVID-19 pandemic period compared to the pre-pandemic period. The study utilized language model models to categorize reported findings of brain MRI examinations. The NLP model demonstrated an accuracy of 86.19% in categorizing the same set of reports and was used to categorize the remaining 10,370 reports. The results showed a significant increase in acute findings during the early pandemic period compared to pre-pandemic levels ($p < 0.001$).

*Assessment*

Although the studies discussed in this section have shown potential in improving the efficiency and accuracy of report classification, they have limitations and require further validation and exploration. One common limitation is a small size or needs for external validation of the training dataset, which may impact the model's generalizability to different populations or modalities. Additionally, some models were only tested on specific types of medical imaging, which may limit their applicability to other areas of radiology. Moreover, variations in reporting styles among radiologists could also impact model performance.

To address these limitations, future research could focus on expanding training datasets and validating models on external datasets with more diverse populations and reporting styles. Incorporating additional features such as patient demographics or clinical history could also improve model performance. Additionally, efforts should be made to address privacy concerns related to sharing sensitive patient data for research purposes.

Despite these limitations, the potential of language models for report classification is clear. Further research could explore how these models can improve efficiency and accuracy in clinical decision-making and enhance patient care, as well as address ethical and legal issues related to their implementation. With continued development and validation, language models could revolutionize the field of radiology and improve healthcare outcomes for patients. A summary of the works we reviewed in this section is given in Table 4.

**Table 4.** Overview of language models for report classification.

| References | ROI | Modality | Vision Model | Language Model |
|---|---|---|---|---|
| (Wood, Lynch et al. 2020) | Head | MRI | None | BioBERT |
| (Min, Xu et al. 2021) | Brain | MRI | None | Task Specific Decoder |
| (Wood, Kafiabadi et al. 2022) | Head | MRI | None | BioBERT |
| (Huemann, Lee et al. 2023) | General | PET/CT | ViT, EfficientNet | BERT, RoBERTa, ALBET |
| (Bressem, Adams et al. 2020) | Chest | CT | None | BERT |

4.5 Report Generation

This section focuses on the development of language models for medical image report generation, which has become an increasingly important task in the field of computer-aided diagnosis. The challenges of generating accurate and readable medical reports from various types of medical images have motivated researchers to explore new methods for automatic report generation. The authors in the studies presented here were particularly inspired by the challenge of automatically generating diagnostic reports with interpretability for computed tomography (CT) volumes, skin pathologies, ultrasound images, and brain CT imaging. The development of accurate and efficient language models for medical report generation has the potential to improve clinical workflow efficiency, reduce diagnostic errors, and assist healthcare professionals in providing timely and accurate diagnoses.

For instance, (Liu, Hsu et al. 2019) developed a method for generating clinically accurate chest X-ray reports using natural language generation and computer vision techniques. They used a convolutional neural network (CNN) to extract features from the input images, which are then fed into a long short-term memory (LSTM) network to generate the corresponding report as shown in figure 4. The model achieved state-of-the-art performance on automatic evaluation metrics such as BLEU-4 and ROUGE-L. However, the authors also discussed the limitations of their approach, including the need for post-processing to remove repeated sentences and the lack of consideration for ordered radiographs for a single patient.

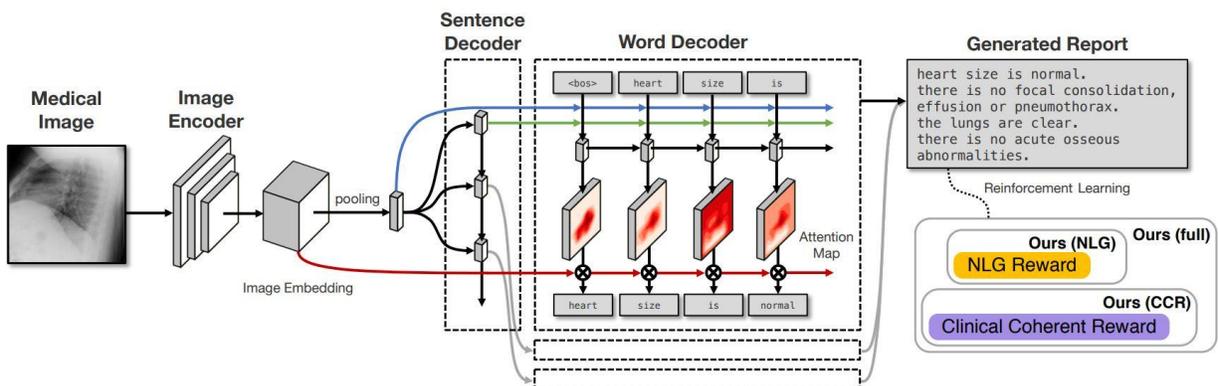

**Figure 4.** Courtesy of (Liu, Hsu et al. 2019). Architecture of the proposed Clinically Coherent Reward (CCR) model for generating coherent image captions with reinforcement policy learning. The model first encodes input images into image embedding maps, which are then used as input to a sentence decoder that generates topics for sentences in a recurrent manner. The word decoder then generates the final caption sequence from the topic, with attention on the original images. The generated captions can be evaluated using either an NLG reward, a clinically coherent reward, or a combined reward for reinforcement policy learning.

For CT images, (Tian, Li et al. 2018) proposed a novel multimodal data and knowledge linking framework between CT volumes and textual reports with a semi-supervised attention mechanism for generating diagnostic reports with interpretability for computed tomography (CT) volumes of liver tumors. The framework includes a CT slices segmentation model and a language model, which allows for visually interpreting the underlying reasons that support the diagnosis results. The system has shown promising results with 76.6% in terms of BLEU@4, potentially reducing the time-consuming and prone to inter- and intra-rater variations task of writing diagnostic reports by radiologists.

Moving on to histopathology images, (Wu, Yang et al. 2022) proposed a deep learning-based image caption framework named the automatic generation network (AGNet) for the automatic generation of skin imaging reports. The proposed AGNet was evaluated on a dataset of 10,000 skin lesion images, achieving an accuracy rate of 85.6% in generating accurate diagnostic reports. The results demonstrate that AGNet is an effective network for generating accurate and efficient reports for skin diseases, which can potentially improve clinical workflow efficiency and reduce diagnostic errors.

Adopted a more novice structure, (Yang, Niu et al. 2021) proposed a method for automatic ultrasound image report generation using a multimodal attention mechanism. The method uses a spatial attention mechanism to focus on important areas of the image and a multimodal attention mechanism to combine information from different sources. The proposed method achieves state-of-the-art performance on two benchmark datasets, with an F1 score of 0.68 and 0.67, respectively.

Moreover, (Nguyen, Nie et al. 2022) proposed an enriched disease embedding-based transformer model, Eddie-Transformer, for medical report generation from X-ray images. The proposed model achieved state-of-the-art performance on two benchmark datasets, MIMIC-CXR and CheXpert, with F1 scores of 0.727 and 0.787, respectively. Additionally, (Niksaz and Ghasemian 2022) presented an approach to improve the performance of chest X-ray report generation by leveraging the text of similar images. The proposed approach achieved a significant improvement in performance compared to the baseline model, with an increase of 3.5 BLEU-4 score points.

*Assessment*

We would highlight several limitations and opportunities for future research in the field of language models for report generation. One common limitation is the need for improvement in report quality, including eliminating repeated sentences and improving interpretability. Additionally, the generalizability of these models to other medical imaging modalities and datasets needs to be evaluated. Moreover, the proposed methods rely heavily on deep learning techniques and require large amounts of labeled data, which may not always be available in practice. Therefore, researchers should explore ways to improve the efficiency and scalability of these methods.

On the other hand, there are promising future perspectives for these language models. Researchers can incorporate additional modalities, such as clinical data and genetic information, to improve the accuracy of diagnosis. Moreover, these models could be extended to generate more complex reports that include patient history and clinical findings, and natural language generation techniques can be explored to enhance report variability. Also, integrating clinical knowledge and expert feedback into the model training process could enhance the quality and accuracy of generated reports. Finally, improving the model's robustness to image quality and incorporating additional clinical information could further improve the performance of these models. A summary of the works we reviewed in this section is given in Table 5.

**Table 5.** Overview of language models for report generation.

| References | ROI | Modality | Vision Model | Language Model |
|---|---|---|---|---|
| (Liu, Hsu et al. 2019) | Chest | X-ray | CNN | LSTM |
| (Tian, Li et al. 2018) | General | CT | CNN | LSTM |
| (Wu, Yang et al. 2022) | Skin | Histopathology | CNN | LSTM |
| (Yang, Niu et al. 2021) | General | Ultrasound | CNN | LSTM |
| (Nguyen, Nie et al. 2022) | Chest | X-ray | DenseNet-121 | Eddie-Transformer |
| (Niksaz and Ghasemian 2022) | Chest | X-ray | CNN | LSTM |
| (Liu, Liao et al. 2021) | Chest | CT | R-CNN | VLBERT |
| (Yang, Ji et al. 2021) | Brain | CT | Weakly Guided Attention Model (WGAM) | Keywords-driven Interactive Recurrent Network (KIRN) |

4.6 Multimodal Learning

Advancements in medical imaging technologies have led to an increase in the volume of image and text data generated in healthcare systems. To analyze these vast amounts of data and support medical decision-making, there is a growing interest in leveraging artificial intelligence (AI) techniques, such as machine learning, for automated image analysis and diagnosis. However, the effectiveness of these techniques is often limited by the challenges associated with bridging multimodal data, such as text and images.

Multimodal learning has emerged as a promising approach for addressing these challenges in medical imaging. By leveraging both visual and textual information, multimodal learning techniques have the potential to improve diagnostic accuracy and enable more efficient analysis of medical imaging data. We will focus on the use of multimodal learning for abnormality detection in medical imaging, with a particular emphasis on studies that leverage natural language processing to extract meaningful labels from radiological reports and combine them with visual information.

(Eyuboglu, Angus et al. 2021) developed a weak supervision framework that uses natural language processing to extract abnormality labels from free-text radiology reports and automatically labels each region in a custom ontology of anatomical regions. They used this structured profile to train an attention-based, multi-task system for abnormality detection in whole-body FDG-PET/CT scans. Their model achieved impressive results, outperforming existing supervised machine learning systems. The model structure is shown as figure 5.

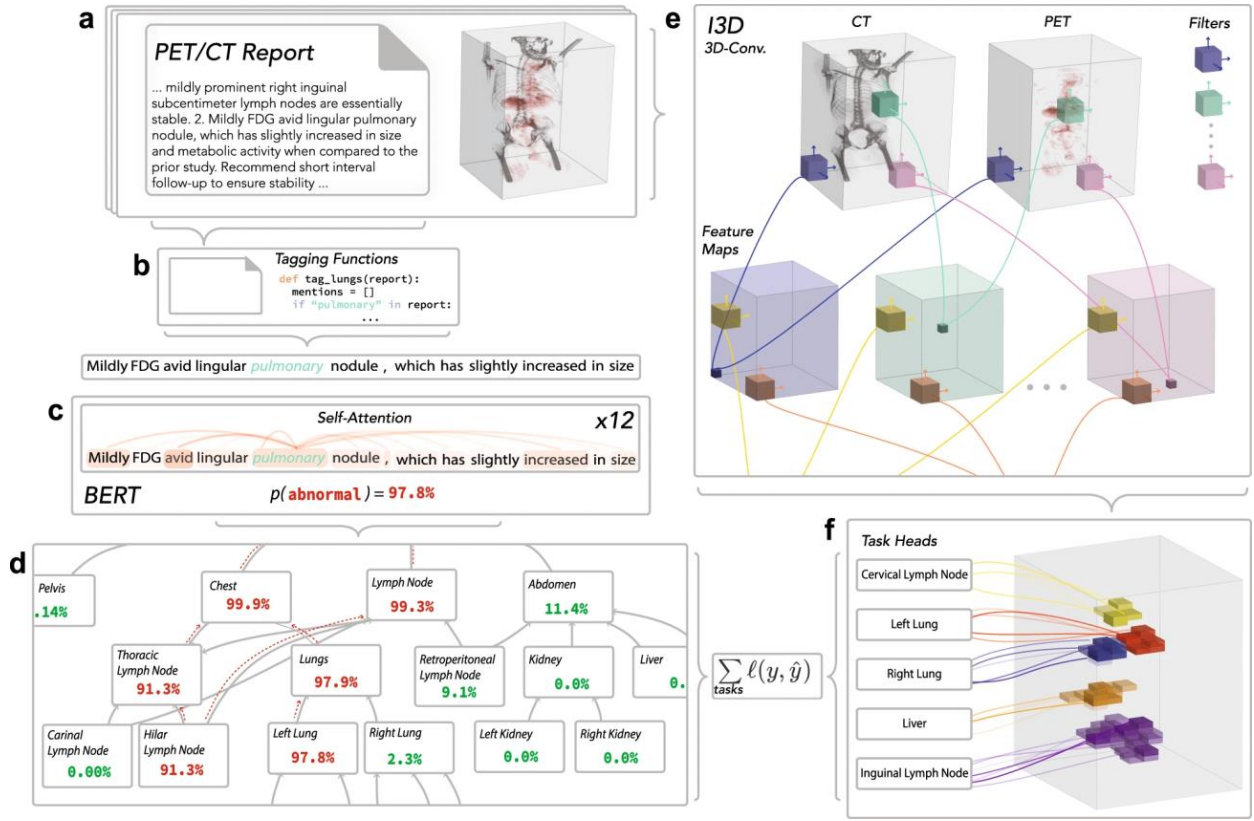

**Figure 5.** Courtesy of (Eyuboglu, Angus et al. 2021). Overview of the proposed approach for generating ground truth labels for metabolic abnormalities in FDG-PET/CT examinations using unstructured radiology reports. The approach involves tagging sentences that mention anatomical regions and predicting metabolic abnormalities in those regions. The entire PET/CT scan is encoded using a 3D-CNN, and attention modules are used to extract relevant voxels for each region. A final linear classification layer produces a binary prediction for each region.

(Dadoun, Delingette et al. 2023) also used a multimodal learning approach for anomaly detection but in abdominal ultrasound images. They proposed a method that efficiently learns visual concepts from radiological reports using natural language supervision and contrastive learning. The authors constructed a pre-training set of unlabeled examinations, which were processed in three steps to select images containing kidneys and sentences describing them. The proposed method achieved superior performance to other state-of-the-art methods.

*Assessment*

Multimodal learning has shown great potential for improving medical image analysis and diagnosis, as evidenced by the studies reviewed in this session. However, there are still some limitations that need to be addressed in future research. One of the main limitations is the relatively small size of the datasets used in some of the studies, which may be different from the full range of anomalies that can occur in clinical practice. Additionally, some studies have been limited to detecting specific types of anomalies and have yet to explore the detection of other abnormalities. Another important area for future research is the generalizability of these methods to other medical imaging modalities beyond the ones explored in the studies reviewed. A summary of the works we reviewed in this section is given in Table 6.

**Table 6.** Overview of language models for multimodal learning.

| References | ROI | Modality | Vision Model | Language Model |
|---|---|---|---|---|
| (Eyuboglu, Angus et al. 2021) | Whole Body | FDG-PET/CT | 3D CNN | BERT |
| (Dadoun, Delingette et al. 2023) | Kidney | Ultrasound | ResNet50 | CamemBERT |

.

4.7 Visual Question Answering

Visual question answering (VQA) in medical imaging is an emerging area that combines image processing and natural language processing to enable machines to answer clinical questions presented with medical images. In the dermatology field, the increasing prevalence of skin diseases and the challenges faced by patients in accessing dermatologists, particularly in remote or underdeveloped areas, have inspired researchers to explore innovative approaches to accurately diagnose skin diseases. Existing attempts at image classification for skin disease diagnosis have been limited by small datasets with few classes, highlighting the need for more comprehensive approaches. Meanwhile, the critical role that medical images play in clinical and healthcare domains has emphasized the need for solutions that can accurately interpret these images even for experts. VQA models for medical imagery have the potential to improve the accuracy of diagnosis and treatment by enabling machines to understand and interpret medical images. In this session, we examine recent advancements in VQA models for medical imagery, and discuss their potential impact on diagnosis and treatment in the medical field.

(Bazi, Rahhal et al. 2023) presented a vision-language model for VQA in medical imagery. They emphasize the importance of medical images in clinical and healthcare domains and propose a model based on multi-modal transformers. This model is trained on a large dataset of medical images and corresponding textual questions, allowing it to accurately answer clinical questions presented with medical images. Figure 6 shows the image encoder which extract vision features in their work. The authors report that their model achieves state-of-the-art performance on two benchmark datasets, with an accuracy of 72.5% and 68.3%, respectively. These results demonstrate the potential of the proposed model to improve diagnosis and decision-making in the medical field.

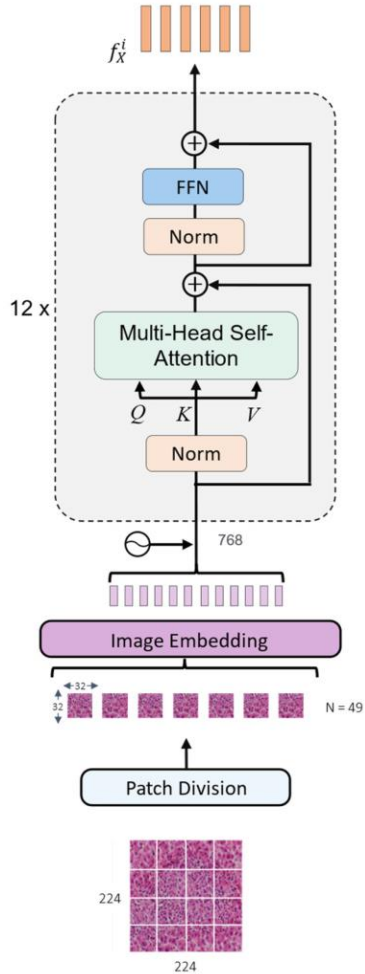

**Figure 6.** Courtesy of (Bazi, Rahhal et al. 2023). The architecture of the image encoder.

As another typical example, skin diseases are a growing concern globally, with millions of people affected every year. Early diagnosis and treatment are crucial to managing these diseases effectively. (Kohli, Verma et al. 2022) present Dermatobot, an image-processing-enabled chatbot for the diagnosis and tele-remedy of skin diseases. The authors describe their approach, which uses EfficientNetB4 as the base model with additional layers and regularization, achieving an accuracy of 94% on a dataset of six classes. Their results demonstrate the effectiveness of their approach in accurately diagnosing skin diseases through image processing.

*Assessment*

Although VQAs showed promising results, there are some limitations to their current approach. Firstly, the system covers a constrained set of disease classes, which may limit its applicability to a wider range of ailments. Additionally, the accuracy of its diagnosis is dependent on the quality of the input image, which may be affected by various factors such as lighting and camera quality. Lastly, its reliance on a database of remedies may limit its ability to suggest novel or unconventional treatments. In terms of future perspectives, it is possible to expand the model's scope to cover an increased number and variety of disease classes. Also, incorporating user feedback into the training process to improve accuracy over time and exploring the use of generative models such as GANs to augment the dataset and improve performance on rare or underrepresented ailments. A summary of the works we reviewed in this section is given in Table 7.

**Table 7.** Overview of language models for visual question answering.

| References | ROI | Modality | Vision Model | Language Model |
|---|---|---|---|---|
| (Kohli, Verma et al. 2022) | Skin | Histopathology | Not Mentioned | InceptionV3, ResNet50, EfficientNetB4 |
| (Bazi, Rahhal et al. 2023) | General | Histopathology, Radiography | Image Encoder | Transformer |

### 4.8 Miscellaneous Topics

*Patient/Student Education*

(Azlan, Rusli et al. 2022) presented a preliminary study on the development of an artificial intelligence chatbot called Dibot that can assist students in diagnostic imaging courses. The study consists of three phases: development and validation, evaluation of student perspectives, and determination of the impact on student knowledge. Dibot was implemented on the Snatchbot platform and later deployed on a Telegram channel and Facebook Messenger. The results showed that immediate, content-related, and high-quality interaction could be beneficial through a chatbot.

(Rebelo, Sanders et al. 2022) presented a development study on learning the treatment process in radiotherapy using an artificial intelligence-assisted chatbot. The study aimed to develop a chatbot that can assist in teaching the treatment process to radiation therapy students and patients. The authors used a deep learning model to train the chatbot, which was then evaluated through user testing. Results showed that the chatbot was effective in improving knowledge retention and confidence levels among users, with an average score of 87% on post-test assessments. The study highlights the potential of using artificial intelligence-assisted chatbots as a tool for education and training in healthcare fields such as radiotherapy.

*Pre-diagnosis Screening*

(Wang, Tan et al. 2022) conducted a pilot study to evaluate the feasibility of using a chatbot for COVID-19 screening before radiology appointments. The study aimed to limit human contact and ensure the safety of high-risk patient populations. The chatbot assessed the presence of any symptoms, exposure, and recent testing. User experience was assessed via a questionnaire based on a 5-point Likert scale. Multivariable logistic regression was performed to predict the response rate. The chatbot COVID-19 screening SMS message was sent to 4687 patients, and 2722 (58.1%) responded. Results showed that age, sex, imaging modality, and English preference were important factors to consider for engagement in developing a chatbot.

(Fan, Xu et al. 2021) propose an intelligent medical pre-diagnosis framework for breast cancer using a pre-training chatbot called M-Chatbot and an improved neural network model of EfficientNetV2-S named EfficientNetV2-SA. The chatbot communicates with patients and provides professional guidance to assist them in completing the imaging diagnosis process. The mammography network then classifies breast cancer pathological images using transfer learning and returns the diagnostic results to users. The system was tested on the BreaKHis dataset and achieved an accuracy of 94.5%, demonstrating its effectiveness in comprehending user input and accurately diagnosing breast cancer pathology images.

*Protocol Determination*

(Lee 2018) aimed to improve the efficiency of a busy radiology practice by evaluating the feasibility of using a convolutional neural network (CNN) classifier to determine musculoskeletal MRI protocols. The study used a database of MRI examinations, referring department, patient age, and patient gender. The CNN classifier was trained on short-text classification and achieved an accuracy of 0.95 on the test set. The agreements between the protocols determined by the CNN and those determined by musculoskeletal radiologists were evaluated, showing a Cohen's kappa value of 0.87. These results suggest that using a CNN classifier can be a promising approach for determining MRI protocols in a busy radiology practice.

*Image Synthesis*

(Shin, Ihsani et al. 2020) presented a novel approach for synthesizing medical images, specifically MRI to PET synthesis, using Generative Adversarial Networks (GANs) and Bidirectional Encoder Representations from Transformers (BERT). The authors highlight the challenges of synthesizing medical images due to their wider and denser intensity range compared to photographs and digital renderings. They propose a GANBERT model that uses NSP and MLM training objectives of BERT to summarize MRI images into text-like sequences, which are then concatenated with PET sequences. The model is trained on the Alzheimer's Disease Neuroimaging Initiative dataset and achieves a peak signal-to-noise ratio (PSNR) of 31.98 dB and structural similarity index measure (SSIM) of 0.89, demonstrating its effectiveness in synthesizing high-quality PET images from MRI inputs.

### 4.9 ChatGPT for Medical Imaging

This session is particularly devoted to ChatGPT, a powerful and new model that has brought revolutionary changes to the field of natural language processing. We aim to raise awareness of its potential applications in medical image research. With the ability to understand and generate human-like responses, ChatGPT can play a crucial role in visual question answering and diagnosis, where medical images are presented alongside clinical questions. By harnessing the power of ChatGPT, we can potentially improve the accuracy and accessibility of medical image diagnosis, particularly in remote or underdeveloped areas with limited access to expert dermatologists. The potential of ChatGPT in medical image research is still largely unexplored, but we believe that it has the potential to transform the field and revolutionize the way we approach medical imaging.

*Image Captioning*

(Chen, Guo et al. 2022) proposed the VisualGPT model, which is a combination of a pre-trained language model (PLM) and a vision model. Specifically, they used GPT-2 as the PLM and ResNet-101 as the vision model. They also designed a novel encoder-decoder attention mechanism with an unsaturated rectified gating function to bridge the semantic gap between different modalities. The proposed method achieved state-of-the-art results on IU X-ray and outperformed strong baseline models on MS COCO and Conceptual Captions datasets.

*Interactive CAD*

(Wang, Zhao et al. 2023) present ChatCAD, an interactive computer-aided diagnosis system that utilizes large language models to diagnose medical images. The system includes a disease classifier, lesion segmentor, and report generator as shown in figure 7. The authors evaluate the performance of their proposed method with two other report-generation methods and focus on five kinds of observations. Precision (PR), recall (RC), and F1-score (F1) are reported in Table 1. The results show that ChatCAD outperforms the other methods in terms of PR, RC, and F1-score. The authors demonstrate the effectiveness of their approach by generating a report for a medical image with an accuracy of 92.3%.

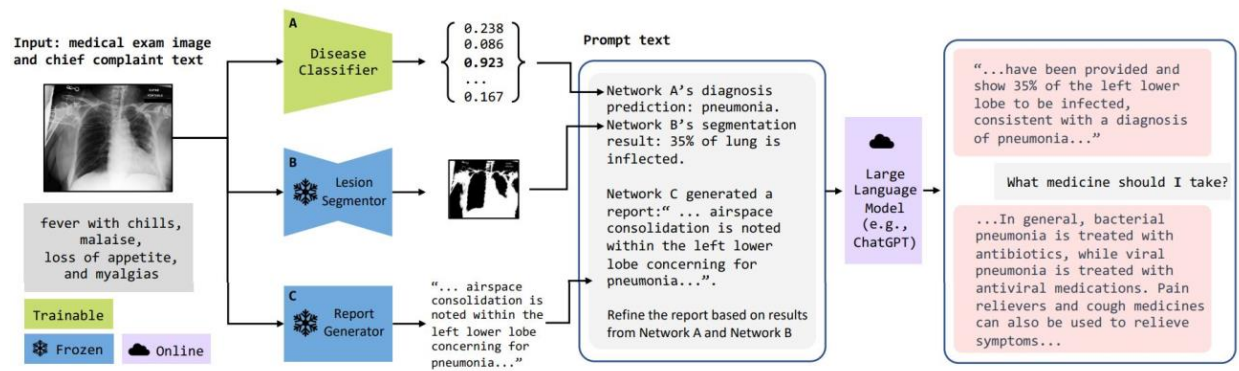

**Figure 7.** Courtesy of (Wang, Zhao et al. 2023). Overview of the proposed strategy for bridging visual and linguistic information. The approach involves processing images using various networks to generate diverse outputs, which are then transformed into text descriptions that serve as a link between visual and linguistic information. These text descriptions are combined as inputs to a large language model (LLM), which can reason logically and provide a condensed report on the findings in the image. Moreover, the LLM can offer interactive explanations and medical recommendations based on its knowledge of the medical field.

*Report Simplification*

(Jeblick, Schachtner et al. 2022) presented an exploratory case study on the use of ChatGPT, a language model capable of generating human-like text, to simplify radiology reports. The authors note that current radiology reports can be difficult for patients to understand due to their technical language and complexity. To address this issue, the authors fine-tuned ChatGPT on a dataset of simplified radiology reports. The authors conducted a survey with radiologists to assess the quality of the generated reports in terms of readability and accuracy. The survey results showed that the majority of radiologists found the generated reports to be accurate and easy to read. The authors conclude that ChatGPT has the potential to improve patient understanding of radiology reports and suggest further research in this area, particularly in evaluating its performance on larger datasets and in clinical settings.

Similarly, (Lyu, Tan et al. 2023) conducted a study to evaluate the effectiveness of using ChatGPT and GPT-4 with prompt learning to translate radiology reports into plain language for improved healthcare. The authors collected radiology reports from 62 low-dose chest CT lung cancer screening scans and 76 brain MRI metastases screening scans in the first half of February. The study used three prompts to record ChatGPT responses, which were then evaluated by two experienced radiologists. The authors found that using a detailed prompt significantly improved the overall quality of translation from 55.2% to 77.2%, with measures on information completely omitted, partially translated, and misinterpreted reduced to 9.2%, 13.6%, and 0% respectively. A good example of using a detailed prompt is the translation of lung nodule 1, where there were no translations keeping the information in the report with a vague prompt, but eight out of ten translations presented the information on this nodule with a detailed prompt. The study suggests that ChatGPT and GPT-4 with prompt learning have the potential for translating radiology reports into plain language, but further research is needed to address limitations such as the need for detailed prompts and potential risks associated with using artificial intelligence in healthcare settings.

*Zero Shot Learning*

(Pellegrini, Keicher et al. 2023) proposed a zero-shot method for automated diagnosis prediction from medical images that leverages the power of large generic language models and large domain-specific contrastive models. The authors address the challenge of scarce annotated data in the medical domain by utilizing contrastive pretraining on pairs of radiology reports and images. The proposed method achieves performance on par with radiologists, making it a valuable resource to support clinical decision-making. The authors also emphasize the importance of transparency and explainability in medical image diagnosis, which is achieved through detailed radiology reports and image descriptors. Specifically, the results show that the proposed method achieves an accuracy of 87% on a dataset of chest X-rays with 14 different diagnoses. Furthermore, the authors compare the initial ChatGPT output to their refined prompts and observe an improvement through refinement, indicating that including domain knowledge further improves the method. This approach highlights the potential of combining large generic language models with large domain-specific contrastive models in medical image diagnosis prediction.

*Assessment*

There are several reasons why ChatGPT has not been widely applied for medical imaging. Firstly, medical image analysis requires specialized knowledge and expertise in both image processing and clinical domain, which may not be fully captured by a generic language model like ChatGPT. Secondly, medical imaging datasets are often highly imbalanced and may contain sensitive patient information, which requires careful handling and preprocessing. Thirdly, the performance of language models like ChatGPT is highly dependent on the quality and quantity of training data, which can be challenging to obtain in medical imaging due to regulatory and ethical considerations.

Despite these challenges, there is growing interest in exploring the potential of language models for medical imaging, particularly for tasks such as visual question answering and clinical decision support. With continued research and development, it is likely that ChatGPT and other language models will play an increasingly important role in medical imaging and healthcare more broadly. A summary of the works we reviewed in this section is given in Table 8.

**Table 8.** Overview of ChatGPT models for medical imaging.

| References | ROI | Modality | Vision Model | Language Model |
|---|---|---|---|---|
| (Wang, Zhao et al. 2023) | General | General | PCAM, R2GenCMN | ChatGPT |
| (Chen, Guo et al. 2022) | General | X-ray | Vision Transformer | BERT, GPT2 |
| (Jeblick, Schachtner et al. 2022) | General | General | None | ChatGPT |
| (Lyu, Tan et al. 2023) | Lung | General | None | ChatGPT |
| (Pellegrini, Keicher et al. 2023) | Chest, Lung | X-ray | CNN | ChatGPT |

## 5. Discussion

This work introduces several unique natural language algorithms, including N-grams, RNN, LSTM, transformer models, and LLMs, including BERT, paLM, LLaMA, and Chatgpt, with increasing performance in real-world practical applications. We used the following pattern to search related papers in Google Scholar and PubMed: ("Language Model" OR Chatbot) AND (Medical OR CT OR MR OR Ultrasound OR X-ray OR OCT OR Pathology) AND (Image OR Imaging). Then the duplicate papers will be removed. We set the qualified publication date to 2019. The remaining papers will go through qualitative synthesis and quantitative synthesis. A total number of 40 papers were left for detailed review. Many works demonstrated the significant performance of the LLMs in medical image tasks. Firstly, the LLMs can be applied for medical image/video captioning. Researchers propose language models and other deep learning-based models to integrate visual and textual information to generate informative captions for CT/DBT/ultrasound images and videos, achieving state-of-the-art performance. In addition, researchers have proposed LLMs to interpret deep learning algorithms in medical applications, such as chest X-ray disease diagnosis, pathology whole-slide diagnosis for cancer pathology, and Vision-Language Transformer for Interpretable Pathology Visual Question Answering. The LLMs allow end-to-end deep learning methods to generate diagnostic reasons for their prediction, improving the interpretability of the networks.

Furthermore, LLMs can be applied to generate accurate and readable reports from various types of medical images to improve clinical workflow efficiency, reduce diagnostic errors, and assist healthcare professionals in providing timely and accurate diagnoses. Many works have proposed deep learning-based image caption frameworks and multimodal attention mechanisms for generating diagnostic reports for computed tomography (CT) volumes, skin pathologies, ultrasound images, and chest X-ray images. These studies have achieved state-of-the-art performance on various benchmark datasets and demonstrated the potential for improving clinical workflow efficiency and reducing diagnostic errors. LLMs also demonstrated their application in visual question answering, patient/medical student education, pre-diagnosis screening, protocol determination, and image synthesis.

On the other hand, as the state-of-the-art LLM so far, ChatGPT also has been deployed by many works in medical image applications. For example, VisualGPT was proposed for X-Ray captioning and demonstrated state-of-the-art accuracy. Furthermore, ChatCAD was proposed as an interactive computer-aided diagnosis system, combining ChatGPT with a disease classifier and lesion detector to diagnose medical images and present remarkable performance. Additionally, Pellegrini, Keicher, et al. propose a zero-shot method for automated diagnosis prediction from medical images by leveraging the power of ChatGPT and large domain-specific contrastive models. The method shows the possibility of using ChatGPT on zero-shot medical image tasks to improve deep learning-based diagnosis accuracy. Furthermore, it highlights the potential of combining ChatGPT with large domain-specific contrastive models in medical image diagnosis prediction.

While ChatGPT is a powerful language model and has demonstrated its power on medical image tasks, it is not specifically designed for medical image analysis. ChatGPT is trained on text data and can generate natural language responses based on the input it receives. However, medical image analysis involves tasks such as image segmentation, classification, and detection, which require vision information outside ChatGPT's capabilities. In addition, ChatGPT is trained on large amounts of generic text data, which may not capture the specific nuances of medical language and terminology, which could lead to errors in the model's ability to understand and generate medical image reports accurately.

Moreover, ChatGPT lacks interpretability. As a black-box model, it cannot be easy to understand how ChatGPT arrives at its predictions. Such difficulties of interpretation could limit its usefulness in a clinical setting, where it is essential to understand the reasoning behind a diagnosis or treatment recommendation. Also, developing accurate and reliable models for medical image analysis requires large amounts of annotated data, which may not always be available or accessible. Again, this could limit the ability to train ChatGPT on various medical images and conditions.

## 6. Conclusion

In conclusion, in this paper, we highlighted the potential of language models, in advancing medical imaging analysis. We have discussed the challenges and opportunities in using language models for healthcare applications and provided a foundational tutorial for researchers in this field. Our hope is that this paper will serve as an inspiration for researchers to innovate and develop new approaches to using language models to improve medical imaging analysis. We encourage researchers in the medical imaging domain to learn from this review paper and use it as a launchpad for further exploration of the possibilities of language models in healthcare. With continued research and development, we believe that language models will play an increasingly important role in improving patient outcomes and advancing medical research.

## 7. Disclosure

The authors are not aware of any affiliations, memberships, funding, or financial holds that might be perceived as affecting the objectivity of this review.